\documentclass[lettersize,journal]{IEEEtran}
\usepackage{amsmath,amsfonts}
\usepackage{algorithmic}
\usepackage{algorithm}
\usepackage{array}
\usepackage[caption=false,font=normalsize,labelfont=sf,textfont=sf]{subfig}
\usepackage{textcomp}
\usepackage{stfloats}
\usepackage{url}
\usepackage{verbatim}
\usepackage{graphicx}

\usepackage{microtype}
\usepackage{cite}
\usepackage{comment}
\usepackage{todonotes}
\usepackage{amsmath,amssymb,amsfonts}
\usepackage{booktabs}
\usepackage{textcomp}
\usepackage{xcolor}
\def\BibTeX{{\rm B\kern-.05em{\sc i\kern-.025em b}\kern-.08em
    T\kern-.1667em\lower.7ex\hbox{E}\kern-.125emX}}
    
\usepackage{paralist}
\usepackage[shortlabels]{enumitem}

\usepackage{booktabs, makecell, tabularx}

\setcellgapes{2pt}
\newcolumntype{L}{>{\raggedright\arraybackslash}X}
\usepackage{caption}
\usepackage{changepage}
\begin{document}

\title{Pick the Right Co-Worker: Online Assessment of Cognitive Ergonomics in Human-Robot Collaborative Assembly}

\author{Marta Lagomarsino,
Marta Lorenzini,
Pietro Balatti,
Elena De Momi,
Arash Ajoudani, \IEEEmembership{Member, IEEE}

\thanks{M. Lagomarsino, M. Lorenzini, P. Balatti and A.Ajoudani are with the Human-Robot Interfaces and Physical Interaction, Istituto Italiano di Tecnologia, Genoa, Italy. 

M. Lagomarsino and E. De Momi are with the Neuroengineering and Medical Robotics Laboratory, Department of Electronics, Information and Bioengineering, Politecnico di Milano, Italy. 

Corresponding author's email: {\tt\small marta.lagomarsino@iit.it}
}
\thanks{Manuscript received February 4, 2022; revised May 2, 2022.}}

\markboth{IEEE Transactions on Cognitive and Developmental Systems,~Vol.~XX, No.~X, Month~2022}%
{Shell \MakeLowercase{\textit{et al.}}: A Sample Article Using IEEEtran.cls for IEEE Journals}


\maketitle

\begin{abstract}
Human-robot collaborative assembly systems enhance the efficiency and productivity of the workplace but may increase the workers' cognitive demand. 
This paper proposes an online and quantitative framework to assess the cognitive workload induced by the interaction with a co-worker, either a human operator or an industrial collaborative robot with different control strategies. 
The approach monitors the operator's attention distribution and upper-body kinematics benefiting from the input images of a low-cost stereo camera and cutting-edge artificial intelligence algorithms (i.e. head pose estimation and skeleton tracking).
Three experimental scenarios with variations in workstation features and interaction modalities were designed to test the performance of our online method against state-of-the-art offline measurements. 
Results proved that our vision-based cognitive load assessment has the potential to be integrated into the new generation of collaborative robotic technologies. The latter would enable human cognitive state monitoring and robot control strategy adaptation for improving human comfort, ergonomics, and trust in automation. 
\end{abstract}


\section{Introduction}
\label{sec:introduction}

\IEEEPARstart{C}{ollaborative} robots have shown their capacity to coexist, cooperate, and safely share the working environment with humans, contributing to better work performances and improving physical ergonomics \cite{Kim2019, Ajoudani2020}. 
Although such collaborative forms of automation provide unique opportunities, they may perilously increase workers' cognitive demand and result in adverse health and safety hazards.
The elevated mental workload, in fact, may affect operators' well-being and consequently compromise their output and the efficiency of the workplace.
Besides, recent surveys \cite{HSEstress2020, Eurofound2019} and systematic reviews \cite{vanderMolene2020} indicate that work-related stress and psychological risks affect to date hundreds of millions of people worldwide, having direct financial implications for private companies and governments \cite{Hassard2014}. 

Consequently, assessing the social acceptance of and trust in collaborative robots in real industrial settings is becoming crucial for developing efficient and effective hybrid manufacturing environments. 
Despite the growing interest in the topic, very little research has been directed towards understanding the effects of automated assistance in the production line and differences were found in how examined groups perceived and responded to the robot. 
In \cite{Wurhofer2015}, workers of a semiconductor factory were interviewed about their interactions with robots over time. Results suggested that the deployment of robots affects workers’ daily activity, but the impact changes acquiring familiarity and experience with the technology and could be reduced by informing the worker as much as possible about the process.
Through subjective questionnaires and narrative interviews, other studies \cite{ElMakrini2018, Sauppe2015} demonstrated that the robot had been accepted as part of the team and that operators manifested their sense of pride in working with the latest technology. 

\begin{figure}[t!]
\centering
\includegraphics[width=\linewidth]{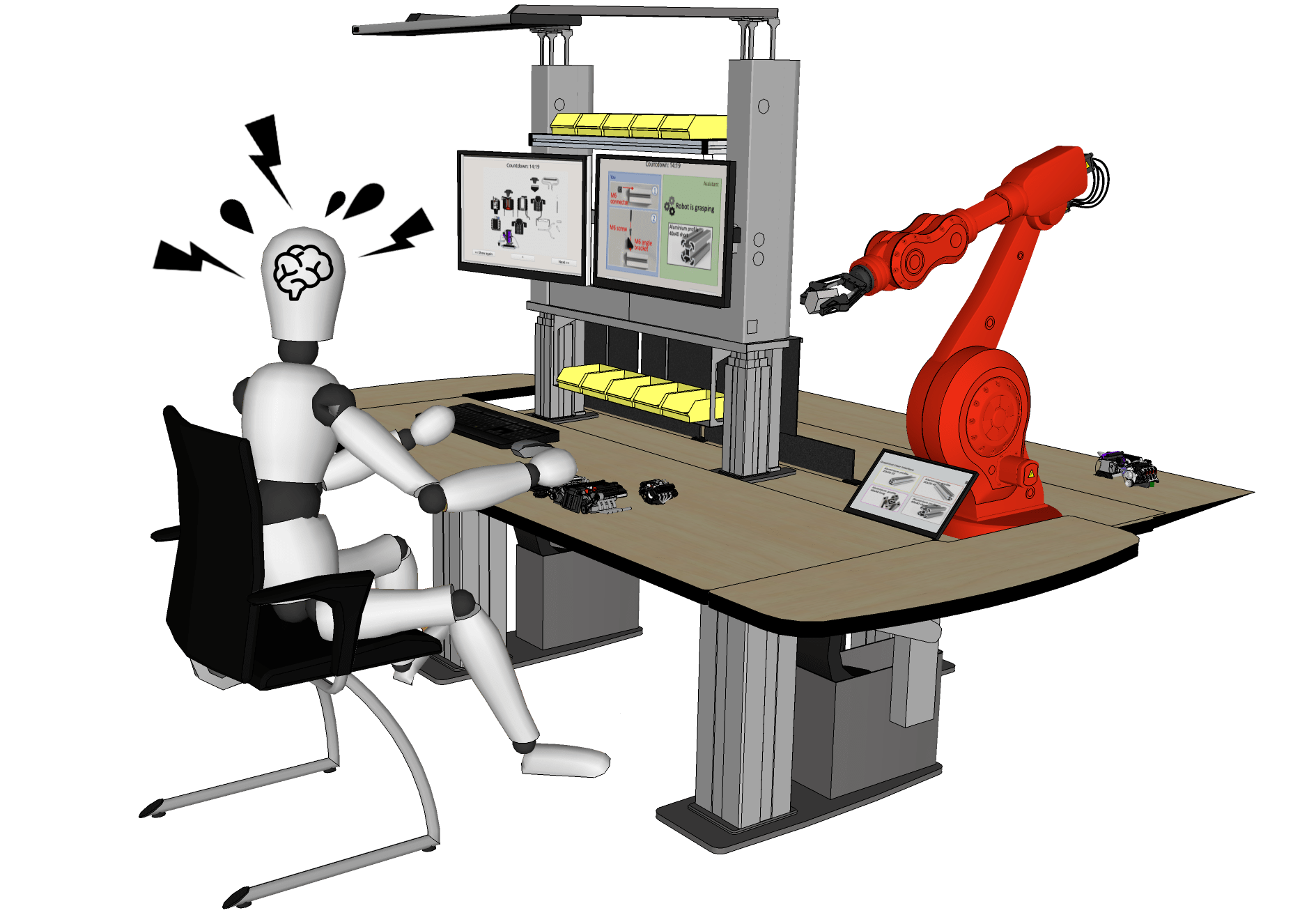}
\caption{
Conceptual illustration of complex assembly operations, potentially increasing worker's cognitive demand, leading to errors and task difficulties. Online and quantitative evaluation of cognitive load, especially in hybrid environments, can safeguard operator's health and ensure efficiency of workplaces.}
\label{fig:system_overview}
\end{figure}

The interrelations between cognitive fatigue, operator sex and robot assistance level were analysed by \cite{Hopko2021} to optimise human-robot interaction and collaboration (HRI-C) system designs with respect to task performance and user experience.
The mental effort, estimated by post-processing the electrocardiography (ECG) signal, negatively affected task efficiency, while the assistance through automation was subjectively perceived and rated in questionnaires as benefitting performance in female subjects. 
Physiological measurements (i.e. ECG, galvanic skin response or GSR, respiratory rate and peripheral skin temperature) were also exploited by \cite{Novak2011psychophysiological} to estimate the mental cost in physically demanding tasks with haptic robots. 
Besides, the analysis of subjective rating scales and secondary-task performance in \cite{kaber2004theeffect} suggested that, if steps of the primary task were automated, worker perceptual resources were freed up and monitoring performance on the secondary task improved.

However, to the best of our knowledge, available tools to model human mental workload and quantify the cost of performing collaborative tasks with a robotic agent can be used almost exclusively by experts or merely provide offline insights into the cognitive process \cite{Thorvald2019, Maurtua2017, Charles2019, Paas2003}. 
A first attempt toward an online cognitive load assessment in industrial HRI-C was made in \cite{Rajavenkatanarayanan2020}. The authors recorded ECG and GSR and utilised machine learning (ML) methods to detect humans' high or low cognitive load while cooperating with a robot. 
Similarly, \cite{Leone2020} exploited a ML classifier to discriminate between stressful and relaxed states through the acquisition of ECG, GSR and electrooculogram signal (EOG) measured by smart glasses and \cite{Salaken2020} tested the feasibility of detecting the cognitive load from the electroencephalographic (EEG) device utilising four classifiers, including random forest, neural network, linear discriminant analysis and logistic regression. 
Nevertheless, such evaluations require rather expensive and impractical equipment and, like most state-of-the-art techniques, it is potentially challenging to be applied in industrial scenarios. 

\begin{figure*}[!t]
\centering
\includegraphics[width=\linewidth]{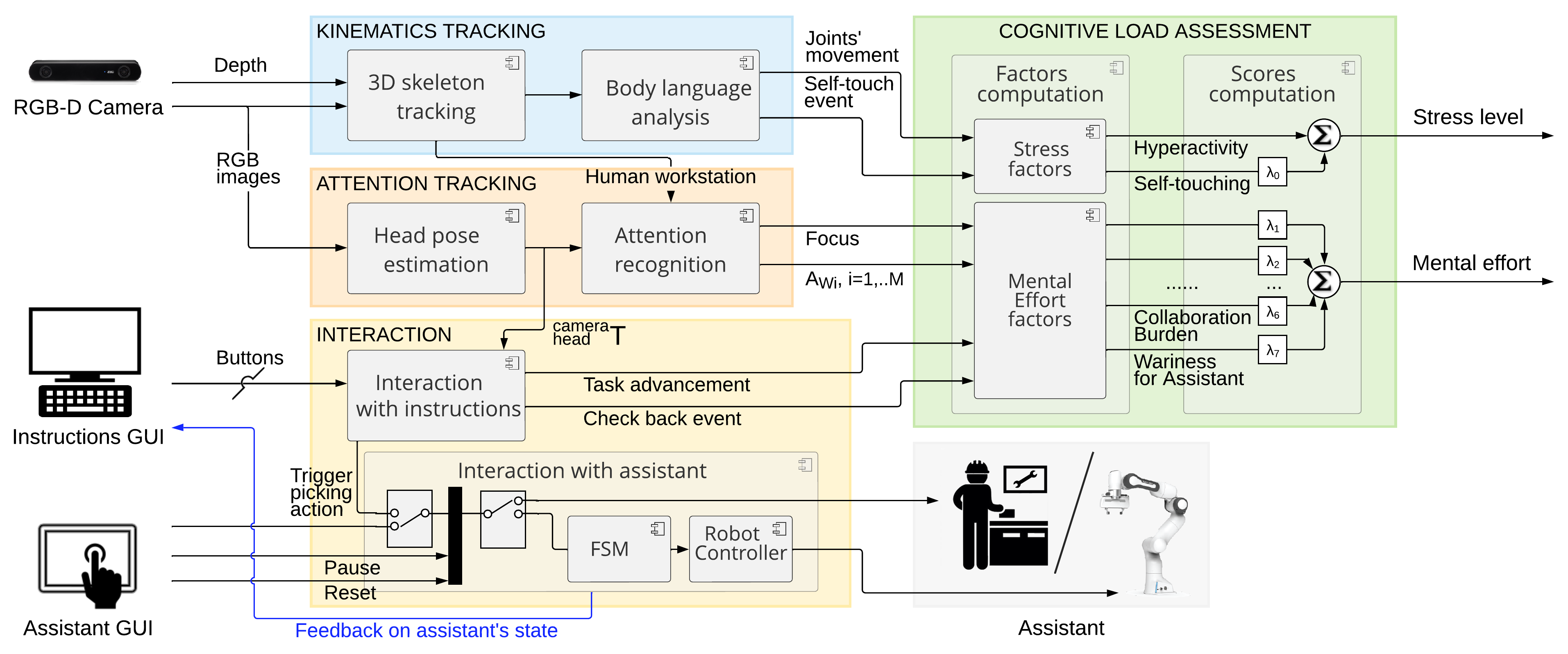}
\vspace{0.01cm}
\caption{Overall structure of the online cognitive load assessment framework detecting patterns in human's motion, investigating workers' attention, their interaction with assembly instructions and trust in the assistant.}
\label{fig:schema}
\end{figure*}

The purpose of this paper is to present an online and quantitative method to appraise the impact of industrial collaborative robots and their actions on operators' cognitive workload in the production line (see conceptual illustration in Fig. \ref{fig:system_overview}).
The proposed framework monitors the mental effort, psychological stress and trust in assistance of human operators directly from the input images of a low-cost RGB-D camera. 
A preliminary study in purely manual assembly tasks with imposed growing complexity is presented and validated in \cite{Lagomarsino2021}. 
Head pose estimation and skeleton tracking are exploited to analyse the workers' attention distribution and assess hyperactivity and unforeseen movements. 
This paper extends the factor assessment tool to consider the interaction with an assistant. Relying on research on gaze tracking and interpretation, we examine the percentage of attention that an individual gives to the assistant during the collaborative assembly task and count the number of glances and gazes over time towards the area dedicated to the assistant.
Moreover, we propose two graphical user interfaces to browse the assembly instructions and handle the interaction with the assistant, and we examine the associated keystroke dynamics.
The study employed assembly experiments in three collaborative scenarios (i.e. human-human and two human-robot settings) with variations in workstation features and interaction modalities to achieve successful acceptance and usage of industrial robotic teammates.
The underlying hypothesis was that high robot assistance level and greater transparency into the robot's autonomous status decreases the cognitive load and increases the trust of the human partner. 
The performance of the developed tool was validated against state-of-the-art methods in the field of cognitive science. ECG and GSR signals were indeed recorded during the task execution and processed offline, along with subjective questionnaires, to ensure and support the validity of the results.

\section{Cognitive Load Assessment Framework}
\label{sec:methods}
The cognitive load assessment framework is illustrated by a block schema in Fig. \ref{fig:schema}. 
Our method analyses
\begin{inparaenum}[(i)]
    \item the attention distribution of a worker from gaze direction and head pose, 
    \item the stress level, by examining activity-related body language (i.e. self-touching occurrences and high activity periods) and 
    \item the cognitive effort required to utilise the assembly instructions and handle the right tools and components to complete the task.
Additionally, we investigate \item the trust in the assistance provided during the industrial task by detecting when the operator turns the head and fixates relevant elements of the collaborator.
\end{inparaenum}
Combing all these factors, we evaluate final scores of \textit{mental effort} and \textit{stress level}, enabling the identification of excessive cognitive load in assembly workers.

Before presenting the functioning of each module, we provide a description of the workstations layout. 
In industrial collaborative assembly, an operating environment\footnotemark\, is defined by at least three types of workstations: 
the \textit{assembly workstation} $W_1$, which is the area occupied by the assembly components, 
the \textit{instructions workstation} $W_2$, providing assembly information and steps to follow through e.g. a monitor, and 
the \textit{assistance workstation} $W_3$, where an assistant (either another human operator or a collaborative robot) is willing to support the assembler and provide necessary components from the storage area.
Based on the number of workstations, our system accordingly associates reference frames in the position specified during a configuration phase. The positions of those reference frames with respect to the operator's head are used to determine the attention level toward every workstation (see Sec. \ref{sec:2B}). 

\footnotetext{ Throughout the paper, the term `operating environment' (or `working area' or `workplace') refers to a place available to manufacturing personnel to carry out work, while `workstation' is a specific location, e.g. an assembly table, where employees perform specific tasks.}

\subsection{Human Upper-Body Kinematics Tracking Module}
\label{sec:2A}
The Human Upper-Body Kinematics Tracking Module detects worker's presence in the operating environment and provides spatio-temporal information about human movements over time.
We exploit an RGB-D camera and adopt the visual skeleton tracking algorithm developed by StereoLabs, to track the human skeleton. The choice of the external sensor system was motivated by the desire to make the proposed framework easily deployable in both laboratory and industrial settings. Noteworthy, the module is scalable to any other person tracking method, such as inertial-based motion-capture systems. 
Among the twenty-five human keypoints (e.g. neck, shoulders, elbows, wrists) extracted in real-time by the algorithm, we select the ones belonging to the upper body, and we analyse their 3D position displacements to compute factors describing the operator \textit{stress level} (see Sec. \ref{sec:2D}).

\subsection{Human Attention Tracking Module}
\label{sec:2B}
The RGB images captured by the stereo camera are also used to detect the human face and identify facial landmarks, exploiting OpenCV library and a TensorFlow pre-trained deep learning model, respectively. Thus, the head pose is continuously estimated by solving a Perspective-n-Point (PnP) problem between the OpenFace 3D model of the face and the output of the detector. A Kalman Filter is included to stabilise the pose computed frame by frame and obtain a more reliable visual-based head tracker. 
Depending on the estimated head position and orientation with respect to the camera, a frame is associated with the worker's head and the transformation $^{\text{camera}}_{\text{head}}T$ represents the head pose variation over time. 
Consequently, the Cartesian vector expressing the relative position between the head and each $i$-th workstation $W_i$ ($i$=$[1,2,\dots,N]$ where $N$ is the number of workstations defined in a configuration phase) is mapped into spherical coordinates (i.e. azimuth angle $\theta_i$, elevation angle $\varphi_i$ and radial distance).

To estimate the level of attention toward each  workstation, we model a fuzzy logic membership function 
\begin{equation}
\label{eq:attention}
\vspace{0.1cm}
\!f(\alpha_i)\! =\!\!
\begin{cases}
    1, 
        & \!\!\! \mbox{\small if \normalsize} \! \left\lvert\alpha_i\right\rvert \leq\alpha_{\text{min},i} \\
    \!\frac{1}{2}\! \Bigl[ 
    1\!-\!\cos{\!\Bigl(
    \dfrac{\left\lvert\alpha_i\right\rvert - \alpha_{\text{min},i}}
    {\alpha_{\text{max},i} - \alpha_{\text{min},i}}
    \pi\Bigr)} \Bigr], 
        & \!\!\! \mbox{\small if \normalsize} \alpha_{\text{min},i}\! <\! \left\lvert\alpha_i\right\rvert\! \leq\!\alpha_{\text{max},i} \\
    0, 
        & \!\!\! \mbox{\small if \normalsize} \! \left\lvert\alpha_i\right\rvert > \alpha_{\text{max},i}
\end{cases}
\vspace{0.1cm}
\end{equation}
where the computed angles $\alpha_i$ at each time instant $t$ (i.e. azimuth $\theta_i(t)$ or elevation $\varphi_i(t)$) are separately transformed using a raised-cosine filter 
\cite{Glover2004}
in a predefined range $[\alpha_{\text{min},i},\alpha_{\text{max},i}]$.
The attention level $A_{\textit{W}i}$ toward the $i$-th workstation is therefore defined as the product between the normalised azimuth and elevation indicators as
\begin{equation}
    A_{\textit{W}i}\bigl(\theta_i(t), \varphi_i(t)\bigl)\, =  f\bigl(\theta_i(t)\bigl) \, f\bigl(\varphi_i(t)\bigl).
\end{equation}
Given the estimated attention to all the workstations, we can assess if the worker is currently distracted or focused on a particular task. To this aim, we check if at least one of the attention parameters is above a predefined threshold, and we determine the workstation that the worker is focused on as the one in which the associated parameter $A_{\textit{W}i}$ is maximum. 

\subsection{Interaction with Instructions and with the Assistant}
\label{sec:2C}
The central role of this module is to handle user's inputs through dedicated Graphic User Interfaces (GUIs). 
In this work, we assume that assembly instructions are shown on a monitor and, thanks to keyboard commands, the worker can browse them to gain knowledge about the following mounting step or go back in instructions. 
Accordingly, the `interaction with instructions' block is in charge of monitoring the task advancement, providing the system with the number of steps of the assembly sequence that the user has already followed and the instruction check backs.

During assembly activities, the worker can ask for support from another human operator or an industrial collaborative robot and co-operate with the assistant through different interaction modalities (see Sec. \ref{sec:3A}).
According to the set modality, the request for a given action by the assistant (e.g. picking a component from the storage area) can take place via interactive buttons displayed on a tablet or can be directly triggered by proceeding with the instructions of the assembly sequence.
Additionally, the user can pause and resume the assistant's action and reset it when a fault in the selection is made. 
Finally, the assistant can provide feedback on the advancement of the served action or the current state (e.g. grasping a given object or resetting the action), which is displayed on the monitor of the assembly instructions.  
As mentioned before, the `interaction with assistant' block is scalable either for a human assistant or an industrial collaborative robot.
In the former case, we assume that another operator is working in a close operating area where additional components are stored and he/she can be informed via a screen whenever the assembler needs a component. 
The integration of a collaborative robot in our cognitive load assessment framework was instead achieved using a high-level Finite State Machine (FSM). The FSM enables continuous transitions between robot states in response to the external inputs from the worker and the task advancement.  
The FSM's initial state is identified by the `Input waiting' primitive in which the robot waits for commands to start its motion. 
When the ``component picking'' action is triggered, the robot passes through a sequence of states (i.e. `Reaching component', `Grasping', `Delivering' and `Handing over', also displayed on the visual feedback screen), enabling the robotic arm to pick the corresponding object, bring it close to the human and finally hand over the item to the human subject. 
Subsequently, the collaborative robot moves back to the homing position, ready to serve new requests. 
The robot action can be stopped at any moment or reset, permitting the part to be repositioned in the initial storage position and then triggering the `Homing' primitive.

\subsection{Cognitive Load Assessment Module}
\label{sec:2D}

\begin{table*}[!t]
\centering
\caption{\centering Definition of \textit{mental effort} and \textit{stress level} factors.}
\label{tab:CLfactors}
\begin{tabular}{p{6.5cm}p{5.5cm}}
    \toprule
    \textbf{Mental effort factors} &  \textbf{Expression} \\
    \midrule
\small{\emph{Concentration Loss}: 
    time that the subject does not explicitly dedicate to the task accomplishment.}
    &
    {$\text{1}-\sum_{w = 1}^{M}\dfrac{\text{[attention time]}_w}{\text{time elapsed}}$} \\
\small{\emph{Learning Delay}: 
    ability to rapidly learning a novel rule from instructions and automaticity in completing assemblies/subtasks.} &  
    {$\dfrac{\text{dwell time on assembly}}{\text{time elapsed}} = \dfrac{\text{[attention time]}_1}{\text{time elapsed}}$} \\
\small{\emph{Concentration Demand}: 
    estimation of the  incidence of attention failures.} &  
    $\sum_{d=1}^{D} \dfrac{\text{[instant of attention loss}]_{d}}{\text{time elapsed}}$ \\
\small{\emph{Instruction Cost}:
    estimation of the general quality of instructions.} &
    $\sum_{c=1}^{C} \dfrac{\text{[instant of not required switch]}_c}{\text{time elapsed}}$ \\
\small{\emph{Task Difficulty}:
    estimation of the required cognitive effort to perform tasks.} &
    $\sum_{b=1}^{B} \dfrac{\text{[instant of instruction check back]}_b}{\text{time elapsed}}$ \\
\small{\emph{Collaboration Burden}:
    attention that the subject gives to the assistant during the collaborative task.} &
    $\dfrac{\text{assistant fixation time}}{\text{time elapsed}} = \dfrac{\text{[attention time]}_3}{\text{time elapsed}}$ \\
\small{\emph{Wariness for Assistant}:
    level of trust of the subject toward the assistant.} &
    $\sum_{a=1}^{A} \dfrac{\text{[instant of assistant check]}_a}{\text{time elapsed}}$ \\
    \bottomrule
\end{tabular}

\medskip
\begin{tabular}{p{6.5cm} p{5.5 cm}}
    \toprule
    \textbf{Stress level factors} &  \textbf{Expression} \\
    \midrule
\small{\emph{Self-touching}: 
    behavioural indicator of stress and anxiety.} &
    $\sum_{s=1}^{S} \dfrac{\text{[instant of self-touching]}_s+60-t}{\text{60}}$ \\
\small{\emph{Hyperactivity\footnotemark}: 
    high activity periods with respect to baseline movements in terms of joint's displacement over time.} &  $m_{k}^{j} = \sum_{l=0}^{\tau-1} \,{d_{k-l, k-l-1}^{j}}$
    
    $\text{if  }  \Delta_k^j=m_{k}^{j}\hspace{-2pt}-\hspace{-2pt}\mu_j>\sigma_{j} \text{  then }$
    \,
    $a_k^j=\dfrac{\Delta_k^j}{\sigma_j}-1$
    \,
    $a_k=\text{min}\Big(\dfrac{1}{N} \sum_{j=1}^{N} a_k^j, 1.0\Big)$
    \\
    \bottomrule
\end{tabular}

\medskip
Note: $M$ is the number of workstations, while $D$, $C$, $B$, $A$, and $S$ are the total occurrences of the\\ corresponding event while working on the task  \\
\end{table*}

\footnotetext{ We define movements $m_k^j$ of $j$-th joint ($j$=$1$,$2$,..$N$) in a time window $\tau$ as the sum of 3D position displacements $d_{k-i, k-i-1}^{\,j}$ within two subsequent frames (where $k$ refers to a system pipeline loop). 
In an initial calibration phase, we compute the mean motion $\mu_1$,..$\mu_N$ of upper body joints and their standard deviation $\sigma_1$,..$\sigma_N$.
During task execution, we periodically compute the deviation $\Delta_k^j$ of each joint from the baseline $\mu_j$ and associate a parameter $a_k^j$ by comparison with the stored $\sigma_j$. A unique descriptor of activity level $a_k$ is determined as the mean over all upper body joints' activities.}

The last module computes a set of cognitive workload indicators starting from the output of the modules described above.
Table \ref{tab:CLfactors} presents a brief definition of these factors and the associated pseudo-formulas that provide hints on how they are calculated.
For a more detailed description of the proposed cognitive load factors and scores, as well as the scientific motivation behind their definitions, see \cite{Lagomarsino2021}.

In this work, we include two additional factors for dealing with the interaction and collaboration with an assistant. 
The first is based upon contemporary psychology's claim that the average dwell time decreases with confidence and expertise on collaborative tasks \cite{Holmqvist2011}. 
Accordingly, we analyse the attention that an individual gives to the assistant during the collaborative assembly task and we define the \textit{Collaboration Burden} factor as the ratio between the time (as sum of durations) the assembler looks at the assistant and the time elapsed since the beginning of the assembly. 
The \textit{Wariness for Assistant} factor instead investigates the trust of the operator toward the assistant and its dynamic attitude. Taking inspiration from research on gaze tracking \cite{Dini2017}, we count the number of glances and gazes over time, namely the transitions of the worker's attention in and out of the area dedicated to the assistant.

Note that we do not expect that a single factor directly reflects human cognitive processing. Our position is that a combination of these factors could provide insights into the human cognitive system.
Each factor $f$ (with $f$=$1$,..$F$) is indeed normalised with a threshold $\tau_f$, when needed, and multiplied by a weight $\lambda_f$. The sum of the weighted metrics determines the final scores of \textit{mental effort} and \textit{stress level}.
Note that values of $\tau_f$ for the proposed factors were defined as the maximum registered value for all subjects who took part in the model validation experiments in \cite{Lagomarsino2021}. On that occasion, we also asked participants to rate the relative importance of factors in determining the experienced workload, exploiting a technique developed in NASA-TLX. 
Given the patterns of choices, we computed the weights that a specific subject would associate with each factor $f$. The mean among all subjects for each factor weight determined the value of $\lambda_f$. For new factors (i.e. trust factors describing checks and fixations of $W_3$), we exploited thresholds and weights associated with corresponding quantities concerning instructions (i.e. factors describing checks and fixations of $W_2$).

\begin{figure*}[t!]
\centering
\includegraphics[width=\linewidth]{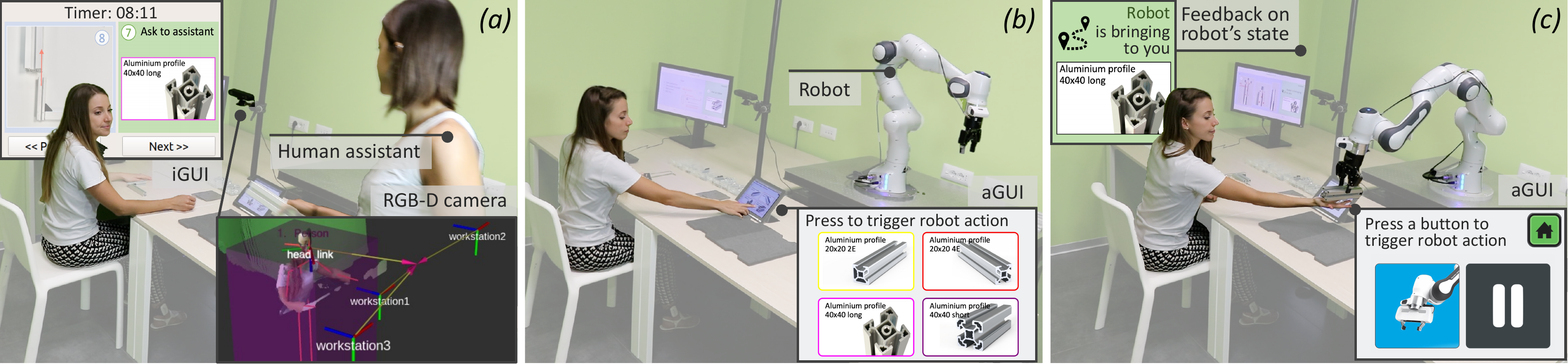}
\caption{Overview of three experimental scenarios: \textit{a)} human-human collaboration, HHC, \textit{b)} human-robot interaction, HRI, \textit{c)} human-robot collaboration, HRC. Employed GUI (i.e. instructions, iGUI, and assistant GUI, aGUI), assistants (i.e. human or robot) and the sensor are also highlighted.}
\vspace{-0.2cm}
\label{fig:scenarios}
\end{figure*}

\section{Experiments}
\label{sec:experiments}
This section presents the experimental campaign to assess how workers’ cognitive load develops while performing a collaborative assembly with an industrial robot. 
The performance of our cognitive load assessment framework was tested against physiological measurements processed after the completion of the experiments. Specifically, we analysed the trend of the \textit{mental effort} in relation to heart rate variability \cite{Hughes2019} and compared the \textit{stress level} with the ordinarily used features in galvanic skin response \cite{Kyriakou2019, Marucci2021}. 

\subsection{Participants}
Fourteen healthy volunteers, seven males and seven females (27.3$\pm$2.9 years old), were recruited for collaborative assembly experiments. The whole experimental procedure was carried out at Human-Robot Interfaces and Physical Interaction (HRII) Lab, Istituto Italiano di Tecnologia (IIT) in accordance with the Declaration of Helsinki, and the protocol was approved by the ethics committee Azienda Sanitaria Locale (ASL) Genovese N.3 (Protocol IIT\_HRII\_ERGOLEAN 156/2020). 

\subsection{Experimental setup}
The participants were asked to sit at a desk, on which aluminium profiles and small boxes with screws, bolts, nuts, etc., were placed (defining workstation $\textit{W}_1$). 
An RGB-D camera (zed2, Stereolabs) monitored the subject from the front for the whole duration of the experiment. 
The assembly instructions were shown on a monitor (workstation $\textit{W}_2$) and could be browsed through a Qt5 C++ GUI, i.e. the instructions GUI (iGUI). 
During the task, an assistant was available to help by providing additional components from the storage area (workstation $\textit{W}_3$). A Python Web Application displayed on a tablet, i.e. the assistant GUI (aGUI), allowed the assembler to select the desired item, as well as pause, resume or reset the assistant's actions.

\subsection{Experimental protocol}
\label{sec:3A}
The study employed a within-subjects experimental design in which each participant underwent three sessions with variations in workstation features and interaction modalities (Fig. \ref{fig:scenarios}). 
In scenario \textit{(a)}, the assembly task was performed in collaboration with another human operator, while in the others (i.e. \textit{(b)} and \textit{(c)}), the assistant was the Franka Emika Panda robot equipped with the Robotiq collaborative gripper.
For all participants, the order of accomplishment of the proposed tasks was randomised and there was a break between sessions to avoid learning effects and cumulative workload. 

Note that we devised an experiment where the assistant provided the operator with five aluminium profiles through handovers distributed during the whole session execution.
Through the scenarios, the team is asked to accomplish different assembly tasks with comparable complexity levels, albeit employing diverse interaction modalities between the assembler and the assistant. HRI-C sessions featured different levels of robot assistance and degrees of transparency into the robot's autonomous status. As such, we expect to identify variations in operators' psycho-social status.

The participants had ten minutes to complete each session, which was estimated as the average time for a seamless task accomplishment.
Before beginning the experiment, a training task was conducted to allow the user to familiarise with the involved interfaces. During this phase, we captured the physiological parameters and upper-body joints movements under resting conditions, which were exploited afterwards as the baseline (see the usage of the mean motion $\mu_j$ and its standard deviation $\sigma_j$ at rest in Tab. \ref{tab:CLfactors}).

\subsection{Collaborative scenarios}
In the following sections, the three proposed collaborative scenarios are described in detail. 

\indent \textit{a)\, Human-human collaboration, HHC:}\, 
The experimental session consists of a collaborative assembly task together with a human assistant. The worker can select the desired item from the aGUI displayed on a tablet and the assistant delivers the requested component. We assume that another individual is working in a close operating area where additional components are stored and he/she can be informed via a screen whenever the assembler needs a component. 
\\[0.4em]
\indent \textit{b)\, Human-robot interaction, HRI:}\, 
The worker interacts with the industrial collaborative manipulator; however, all robot movements must be authorised by pressing suitable buttons on the aGUI. 
Moreover, the robot neither informs the user on the receiving of the part request nor provides any feedback about its state. 
Hence, the degree of interaction is extremely low, and there is no transparency about what the robot is doing.
\\[0.4em]
\indent \textit{c)\, Human-robot collaboration, HRC:}\, 
The robotic agent can perform more tasks in autonomy, but it gives feedback on the monitor about the action currently served (e.g. grasping an object). Robot's actions are directly triggered by proceeding in the assembly sequence through the iGUI. This allows for parallelism in activities performed by the two agents and speeds up the process. Moreover, supervising options (i.e. pausing/resuming robot motion and resetting actions in case of errors) are explained in detail, and their effectiveness is demonstrated to participants before the beginning of the session. 

\subsection{Experimental hypothesis}
\label{subsec:exphp}
The underlying hypothesis was that high robot assistance level and greater transparency into the robot’s autonomous status decreases the cognitive load and increases the trust of the human partner  \cite{Chen2014, Hopko2021}. 
Therefore, investigating workers’ attention toward the assistant, instructions and assembly components and analysing human body language, we expected higher values in our \textit{mental effort} and \textit{stress level} scores in scenario $(b)$ than $(c)$ enabling robot’s feedback. Besides, it would be extremely interesting to analyse differences from human-human trials and access framework performance through questionnaires and physiological measures.  

\subsection{Baseline measurements}
In this section, we present the adopted quantitative and qualitative measures to assess the performance and potential of the proposed framework. Additionally, we justify the choice of specific ground truth parameters and describe the sensors adopted and the post-processing of the acquired signals. 

\vspace{0.1cm}
\subsubsection{HRV responses}

A chest strap (H10, Polar Electro Oy, Kempele, Finland) was used to record the ECG signal. 
A large and growing body of literature \cite{Hughes2019} has indeed investigated the relationship between human cognitive processing and heart rate variability (HRV) metrics. 
Among several metrics identified by researchers, the low-to-high frequency (LF/HF) ratio was selected since it was identified as a biomarker of the mental effort \cite{Durantin2014, Mizuno2011}.
In this work, the raw ECG was initially processed to extract the RR intervals, i.e. the time elapsed between two successive R-waves, and then analysed using Kubios software. 

\vspace{0.1cm}
\subsubsection{Galvanic skin responses}

The galvanic skin response (GSR, also known as electrodermal activity, EDA), a widely studied biomarker of stress,   \cite{Setz2010, Kyriakou2019}, was monitored by the movisens EdaMove4 scientific research instrument.
The mobile device was connected to a textile band worn on the ankle. 
The recorded GSR signal was then processed using the open-source MATLAB toolbox Ledalab. 
A Butterworth low pass filter with a cut-off frequency at 2 Hz was used to filter the high-frequency components.
Finally, we applied the continuous decomposition analysis to separate the tonic (Skin Conductance Level, SCL) and phasic (Skin Conductance Response, SCR) components. 
As in \cite{Marucci2021}, we computed the mean value of the SCL and the mean amplitude of the SCR peaks to investigate the stress induced by the entire task on participants. 

\subsubsection{Subjective questionnaires}

At the end of the experiment, we asked participants to fill NASA task load index questionnaire \cite{Hart1988}, 
to quantify the workload and the trust scale defined in \cite{Charalambous2016}\footnote{Note that the word `robot' was replaced by `assistant' in each statement to evaluate the trust in all three scenarios. Statements $F$ and $J$ were neglected since they were not applicable in our context.} 
to assess trust development in industrial HRI-C. 
Additionally, we asked participants if they had previously performed an experiment in direct interaction with an industrial robot. 
This permitted us to perform a statistical comparison and determine whether familiarity with robotic technologies could reduce the risk of excessive cognitive load.

\section{Experimental results} 

\begin{figure}[!b]
    \vspace{-0.5cm}
    \hspace{-0.5cm}
    \includegraphics[width=1.1\linewidth]{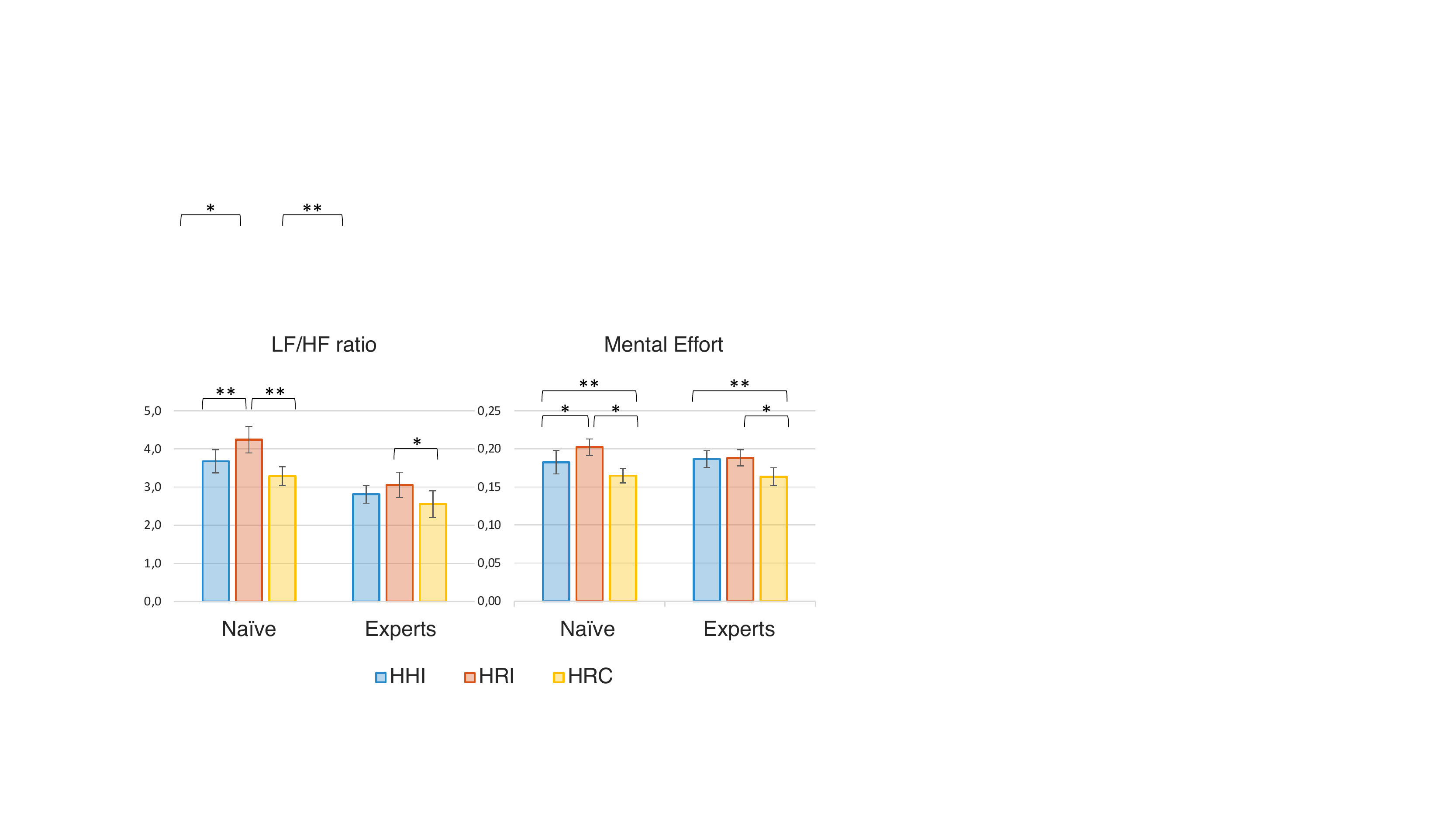}
    \vspace{-6mm}
    \caption{Bar chart indicating mean and standard error of LF/HF ratio and \textit{mental effort} in three scenarios.
    \small Significance levels obtained from Wilcoxon's test are indicated at *p\textless{}0.05, **p\textless{}0.01.}
    \label{fig:histogram}
\end{figure}

\label{sec:results}
In this section, the results of the experimental campaign are presented. 
First, the outcomes of physiological measures and subjective questionnaires are evaluated to determine how different collaborative assembly scenarios have impacted operators' cognitive load. This is followed by a deep analysis of the presented cognitive load scores and their correlations with the ground truth measurements. 
For the study, participants were divided into two groups according to their declared familiarity with robotic technologies. Eight out of fourteen subjects, constituting \textit{group N}, claimed to be na$\ddot{\i}$ve to human-robot collaboration, while the remaining six subjects, \textit{group E}, owned previous experience with robotic manipulators.
Last, we focus specifically on the factors addressing the interaction with the assistant, investigating the differences among the examined interaction modalities and making a comparison with the trust scale results.  

\subsection{Baseline measurements}

\subsubsection{HRV responses}

The ECG signal registered during every collaborative scenario was segmented into four blocks (2.5 minutes each). This permitted us to extract frequency-domain HRV-features within each block and assess differences in the trends using Friedman test with repeated measures.
Different experimental conditions significantly impacted the LF/HF ratio (p=0.022)\footnote{ The p-value obtained from Friedman test is defined as p, while $\text{p}_{i,j}^G$ refers to the p-value resulted from Wilcoxon test with Bonferroni correction between condition $i$ and $j$ ($i,j$=$a,b,c$) for subjects belonging to \textit{group G} ($G$=$N,E$). Finally, median rank of the analysed metric in experimental session $i$ for \textit{group G} is reported as $\text{M}_i^G$.}, a well-known biomarker of the mental effort. 
Figure \ref{fig:histogram} shows the results of the Wilcoxon test with Bonferroni correction. For na$\ddot{\i}$ve subjects, the mean ranks in low human-robot interaction (scenario $b$) were statistically significantly higher than the mean ranks in human-human collaboration session (scenario $a$, $\text{p}_{a,b}^N$=0.007). However, increasing the transparency on robot's actions (scenario $c$), the parameter exhibited a predominant decrease ($\text{p}_{b,c}^N<$0.001). 
On the contrary, the robot agent does not affect the LF/HF ratio for experts ($\text{p}_{a,b}^E$=0.086), and the parameter further decreases in scenario $c$ ($\text{p}_{b,c}^E$=0.033).  

\begin{figure}[!b]
    {\begin{adjustwidth}{-0.7cm}{-0.45cm}
    \includegraphics[width=\linewidth]{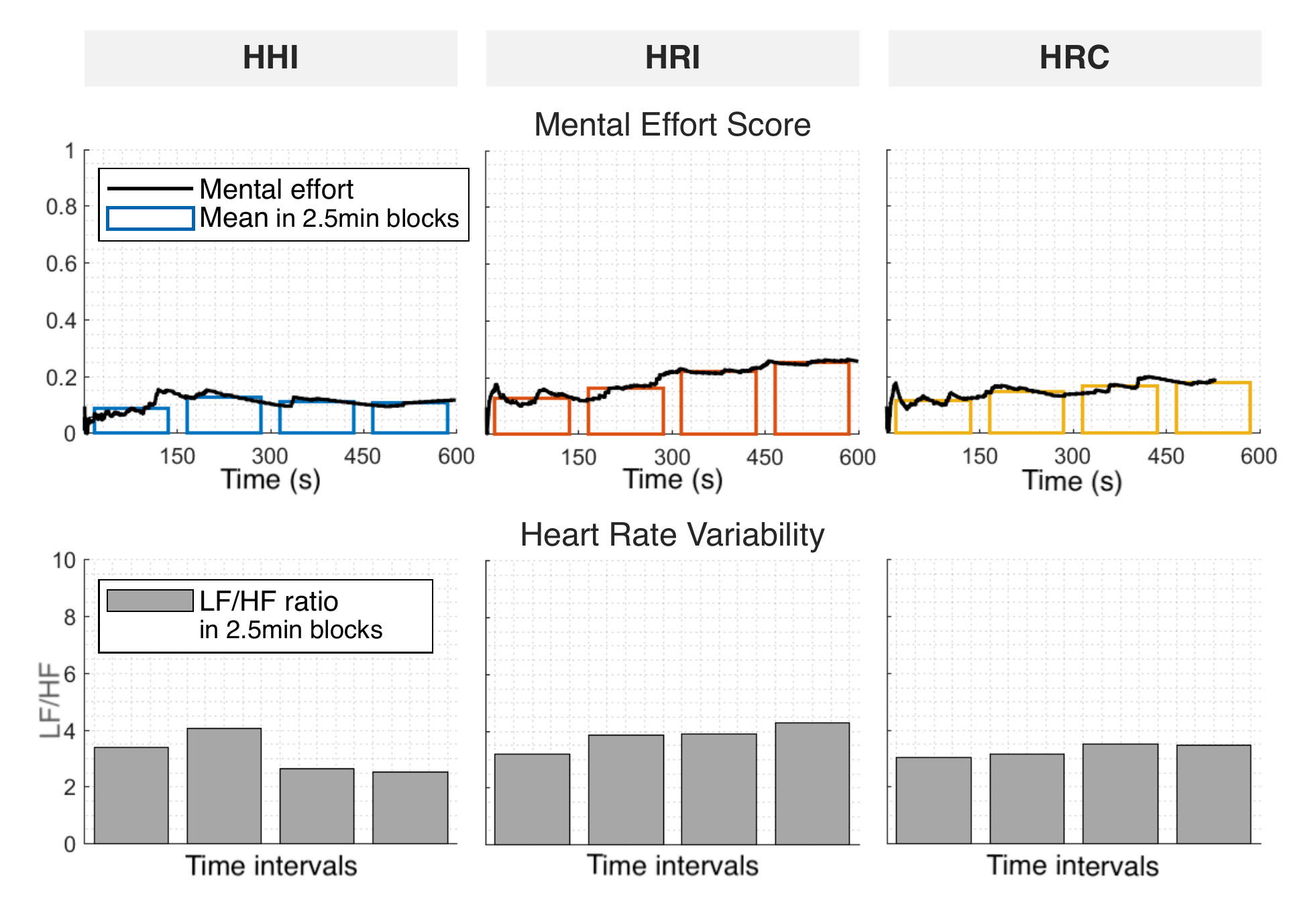}
    \end{adjustwidth}}
    \caption{Comparison between \textit{mental effort} score computed by our online framework and LF/HF ratio extracted from 2.5-minute blocks of electrocardiography signal for a na$\ddot{\i}$ve subject.}
    \label{fig:mental_effort_vs_HRV}
\end{figure}

\subsubsection{Galvanic skin responses}

As in \cite{Marucci2021}, we examined the mean value of the SCL and the mean amplitude of the SCR peaks in different experimental conditions. 
The Friedman test, run in 2.5-minute intervals, revealed a significant main effect of the interaction modalities on the tonic and phasic components (p$<$0.001 and p=0.003, respectively). 
The robotic aid predominately increased the SCL-related parameter compared to the human aid either in na$\ddot{\i}$ve ($\text{p}_{a,b}^N$=0.009, $\text{p}^N_{1,3}$=0.017) or expert ($\text{p}_{a,b}^E$,\,$\text{p}_{a,c}^E<$0.001) subjects. Pairwise significant differences between human-robot interaction were also found in SCL within \textit{group N} ($\text{p}_{b,c}^N$=0.048) and in SCR in both groups ($\text{p}_{b,c}^N$=0.005, $\text{p}_{b,c}^E$0.006).

\subsubsection{Subjective questionnaires}

Median workload levels of NASA-TLX score for scenarios $a$, $b$, and $c$ were 56.7, 52.5, and 40.3, respectively. 
Besides, we computed the total score of trust perceived by participants during the tasks, following the guidelines provided in \cite{Charalambous2016}. 
The Kruskal-Wallis test revealed a significant difference in the score depending on the imposed experimental conditions, p=0.004.
Specifically, na$\ddot{\i}$ve participants reported a significant trust diminution with the robot's introduction ($\text{p}_{a,b}^N$=0.031, $\text{M}_a^N$=34.5, $\text{M}_b^N$=24.5), 
but they regained the confidence to robot assistance with system state's observability ($\text{p}_{a,c}^N$=0.125). 
For experts, the reliability of automated assistance was comparable ($\text{p}_{a,b}^E$=0.234) 
or even enhanced as against human assistance ($\text{p}_{b,c}^E$=0.063, $\text{M}_a^E$=29.0, $\text{M}_c^E$=33.5).

\subsection{Cognitive load scores assessment}

\begin{figure}[!b]
    {\begin{adjustwidth}{-0.45cm}{-0.7cm}
    \includegraphics[width=\linewidth]{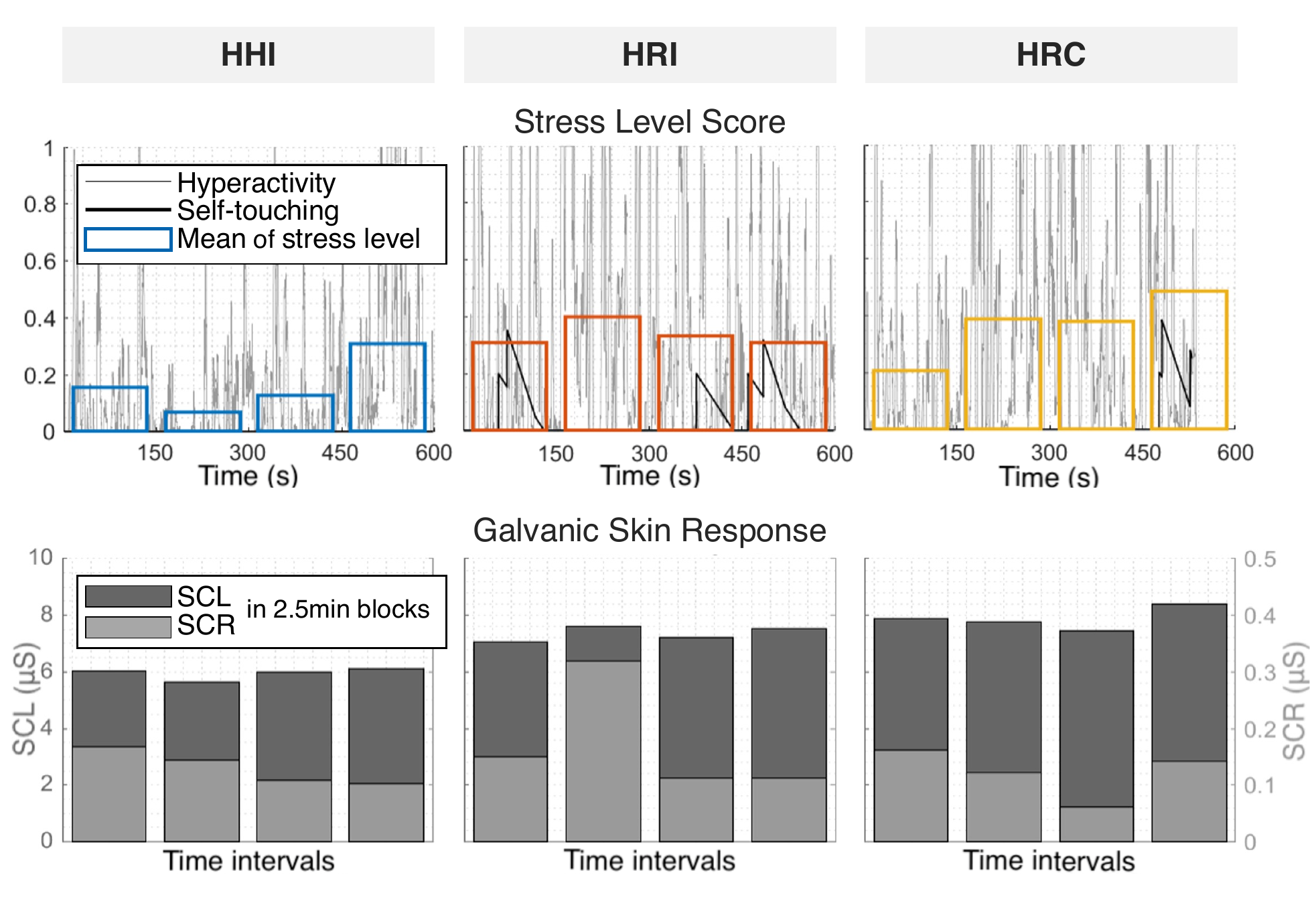}
    \end{adjustwidth}}
    \caption{Comparison between \textit{stress level} score and skin conductance level (SCL) and response (SCR) extracted from 2.5-minute blocks of galvanic skin response for a na$\ddot{\i}$ve subject.}
    \label{fig:stress_vs_GSR}
\end{figure}

\subsubsection{Mental effort}

The \textit{mental effort} score, obtained from the weighted combination of the factors, is presented in Fig. \ref{fig:mental_effort_vs_HRV} (first row) for a na$\ddot{\i}$ve subject. The black line defines the score's trend over time in the different experimental sessions, while coloured bars highlight the mean within 2.5-minute intervals. Note that the participant completed scenario \textit{(c)} before the total available time (i.e. ten minutes).

Analysing all subjects, Friedman test indicated that different interaction modalities affected the \textit{mental effort} score significantly (p=0.004). 
Post-hoc comparisons reveal no differences between human-human and human-robot interaction with limited transparency on robot's actions for experts. 
A significant decrease in the score was however manifested between the two different human-robot interaction modalities ($\text{p}_{b,c}^E$=0.007). 
In na$\ddot{\i}$ve subjects, the \textit{mental effort} score was significantly affected by the robot's presence ($\text{p}_{a,b}^N$=0.043) but a greater transparency on robot's action determined a predominant reduction ($\text{p}_{b,c}^N$=0.054). 

The similarity of statistical outcomes encouraged us to compare the \textit{mental effort} score computed online by our framework and with the offline-extracted LF/HF ratio levels (see Fig. \ref{fig:mental_effort_vs_HRV}).
Positive correlations (Spearman's rank-order correlation coefficient $r_s>$0.4) were found between the mean within the defined blocks of our score and the HRV feature extracted in the same interval for thirteen out of fourteen participants, of which five were significant at 1\% level and two at 5\% level. 

\subsubsection{Stress level}

Figure \ref{fig:stress_vs_GSR} (first row) illustrates the estimated stress of a na$\ddot{\i}$ve participant during the experiments. 
\textit{Hyperactivity} and \textit{self-touching} factors are set out by grey and black profiles, respectively. By summing them, we evaluated the \textit{stress level} score and its mean within blocks lasting 2.5 minutes (displayed through coloured bars). 
Statistically significant differences were identified by the repeated measure Friedman test (p$<$0.001).
Overall, we identified higher score readings during robot assistance versus human assistance. All pairwise post-hoc comparisons showed p-values below the significance level. 
This guided us to analyse the \textit{stress level} trend in relation to the EDA recording (see Fig. \ref{fig:stress_vs_GSR}).
The mean of our score in the defined blocks appeared to be positively correlated to the SCL feature for ten subjects and to the SCR feature for three more subjects, with significance in five subjects.

\subsection{Trust in robot assistant}

\begin{figure}[!t]
\centering
    \includegraphics[width=0.75\linewidth]{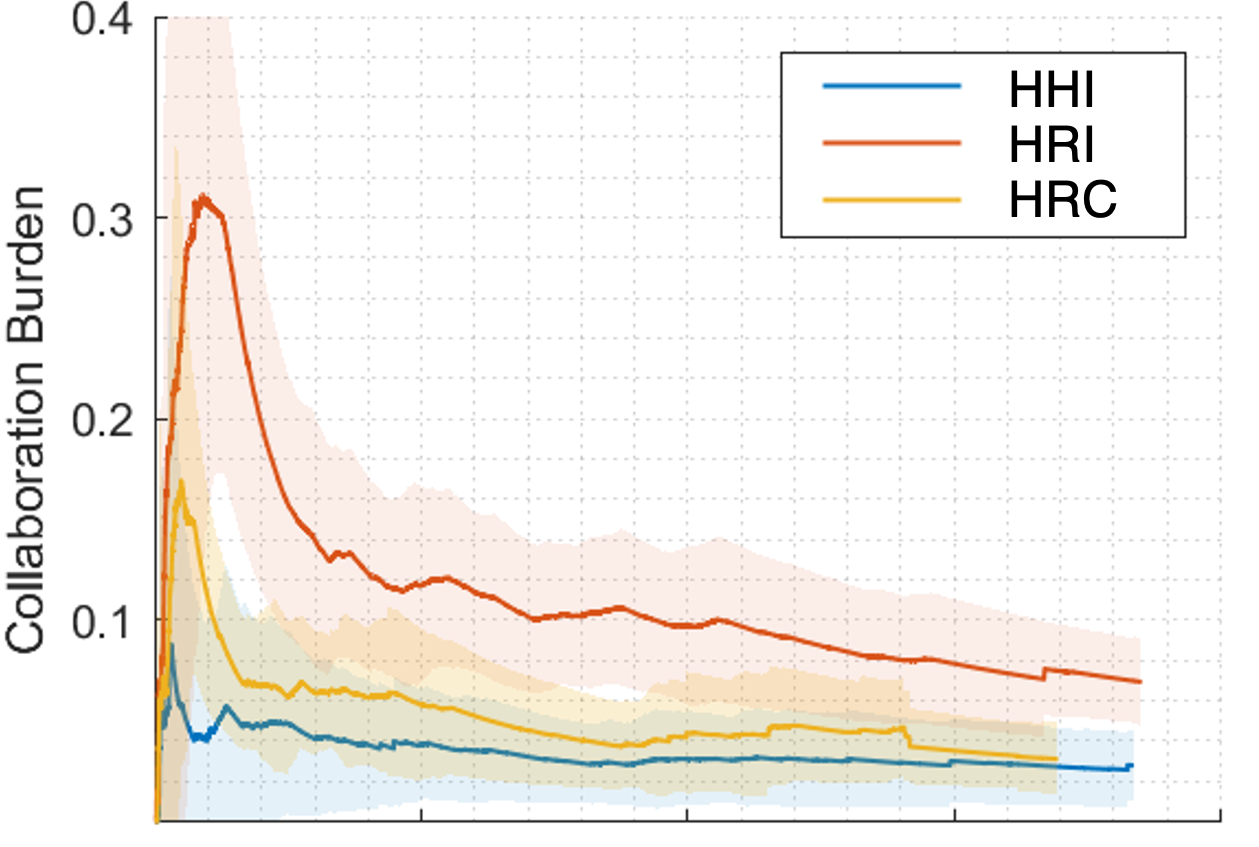}
    
    \hspace{0.1cm}
    \includegraphics[width=0.75\linewidth]{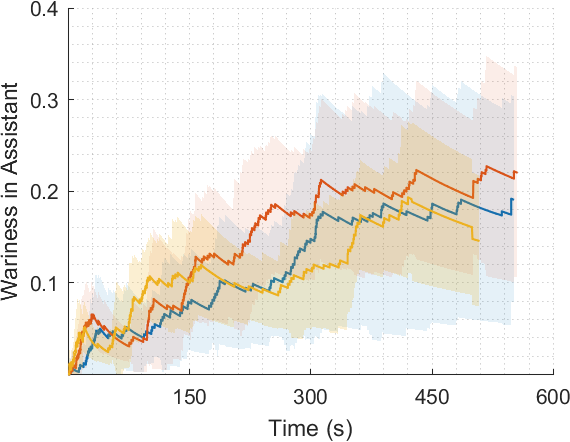}
    \vspace{0.1cm}
    \caption{Mean trend over participants and shaded standard deviation of factors estimating workers' trust in automated and human assistance.}
    \label{fig:trust_factors}
\end{figure}

Figure \ref{fig:trust_factors} displays the developed metrics for operator's trust toward human and automated assistant. 
Note that our factors were defined with pejorative connotation, so $0.0$ refers to complete trust and $1.0$ to loss of any confidence.
We computed the factors mean for all subjects over time (colour-coded line) and the corresponding standard deviation at each system pipeline loop (shaded area). The results for sessions \textit{(a), \textit{(b)}}, and \textit{(c)} are reported on the same chart to highlight differences in the trends. Some participants completed the task before the total available time. Therefore, we stop plotting the mean trend when at least one subject has finished the execution. 

What stands out in the data is an increase of the 7.8\% in the \textit{Collaboration Burden} factor between scenario \textit{(a)} and \textit{(b)}. 
However, the time dedicated by the assembler to the robot reduces with greater transparency on the robot's actions. 
The differences between the three experimental conditions were also found to be statistically significant through the Kruskal-Wallis test (p$<$0.001).
Post-hoc comparisons for na$\ddot{\i}$ve subjects identified a significant growth in the factor between human-human and low human-robot interaction ($\text{p}_{a,b}^N$=0.031), which however vanishes enabling system's transparency ($\text{p}_{a,c}^N$=0.148). 
Interestingly, we also noticed that participants with prior experience in robotics tended to focus 5.1\% less on the robot than the na$\ddot{\i}$ve ones. 
The trends of the \textit{Wariness in Assistant} factor's mean for all participants appeared comparable in scenarios $a$ and $c$ and slightly shifted to higher values in $b$. 
The p-values of Kruskal-Wallis test were, however, over 0.05 significance level.

A Spearman's rank-order correlation was run to determine if a relationship exists between our trust factors and the outcomes of trust scales. For comparisons, we considered the complements of our factors' mean due to the negative sense in which factors are defined in the sessions (e.g. [$1-\mu_{a}^s$, $1-\mu_{b}^s$, $1-\mu_{c}^s$], where $\mu_{i}^s$ is the mean of a factor in task $i$ for $s$-th subject). 
A positive correlation was found in nine out of fourteen for both \textit{Collaboration Burden} and \textit{Wariness in Assistant}. 

\section{Discussion}
The offline statistical analysis on baseline measurements supported our experimental hypotheses (see Sec. \ref{subsec:exphp}) and suggested that the degree of robot's transparency and observability available to the human worker has a noticeable impact on the development of cognitive workload. 
With increasing feedback on the robot's autonomous status, HRV and GSR features indeed indicated a significant reduction of mental effort and stress during the interaction.

Interestingly, changes in experienced cognitive effort between human and automated assistance mainly depended on the operator's familiarity with the technology. 
No differences were indeed found either in ground-truth measures or our cognitive load scores (i.e. \textit{mental effort} and \textit{stress level}) for participants with prior experience in robotics. On the other hand, na$\ddot{\i}$ve subjects tended to see automation as more stressful and cognitively demanding. 

One of the study's major findings was the similarity in the trends of the scores computed online by our method and the state-of-the-art offline measurements that are less likely deployable in industrial settings.
The \textit{mental effort} mean in defined time blocks was positively correlated to the LF/HF ratio derived by ECG signal within the same time intervals. Positive correlations were also identified between the \textit{stress level} and GSR features.

Additionally, we presented and analysed two factors to examine the trust in industrial human-robot collaboration. 
Proposed trust metrics, whose readings were positively correlated to outcomes of a state-of-the-art subjective trust scale, suggested that robotics experts tend to rely more on automated support. However, informed robot movements could put the na$\ddot{\i}$ve human co-worker at ease and foster trust. 

\section{Conclusions}
This study investigated how workers' cognitive load develops while interacting with industrial collaborative robots. 
We proposed an online and quantitative framework to monitor the mental effort and psychological stress of a human operator during an assembly task. 
Attention distribution, high activity periods and body language were extracted directly from the input images of a stereo camera and analysed.
Additionally, two factors were designed to examine the trust in the robotic counterparts within hybrid manufacturing environments. 

The proposed framework works online, does not require expensive equipment and does not ask the human worker to wear any sensor allowing the natural flow of work activities. 
These are promising features to integrate the technique in real industrial settings.
Nonetheless, optical systems induce visibility issues and workers' privacy threats that could be partially overcome with a multi-camera setup exploiting stick figures of the human body, including information about human attention direction. 
Regrettably, the current study was conducted in our laboratory, reproducing a well-structured working environment and was limited to students and staff participants.
Its natural progression will deal with the framework generalisation to complicated industrial operations, examining the optimal camera placement. The study would involve people working in the manufacturing domain and investigate the opportunities for human-robot collaboration in their environment.

This research lays the foundation for future work on cognitive ergonomics in industrial human-robot collaboration. The \textit{mental effort} and \textit{stress level} scores computed online by our framework may be provided as input to the robot enabling real-time adaptation of the control strategy according to human distress and needs. This aspect could be extremely important to improve operators' comfort at work and achieve successful acceptance and use of industrial robotic teammates. 

\section*{Acknowledgment}
This work was supported in part by the ERC-StG Ergo-Lean (Grant Agreement No. 850932), in part by the European Union’s Horizon 2020 research and innovation programme SOPHIA (Grant Agreement No. 871237).

\bibliographystyle{ieeetr}

\bibliography{bibliography}

\vfill

\end{document}